%% file: arxiv_main.tex
\documentclass[acmtog, nonacm]{acmart}
\setcopyright{none}
\renewcommand\footnotetextcopyrightpermission[1]{}
\usepackage{booktabs} 

\usepackage[ruled]{algorithm2e} 

\SetAlFnt{\small}
\SetAlCapFnt{\small}
\SetAlCapNameFnt{\small}
\SetAlCapHSkip{0pt}

\authorsaddresses{}

\begin{document}
\newcommand{\todo}[1]{\textcolor{cyan}{TODO: #1}}
\newcommand{\bfa}{\mathbf{a}}
\newcommand{\bfc}{\mathbf{c}}
\newcommand{\bfd}{\mathbf{d}}
\newcommand{\bfo}{\mathbf{o}}
\newcommand{\bfp}{\mathbf{p}}
\newcommand{\bfr}{\mathbf{r}}
\newcommand{\bfx}{\mathbf{x}}

\title{ƒNeRF: High Quality Radiance Fields from Practical Cameras}

\author{Yi Hua}
\author{Christoph Lassner}
\author{Carsten Stoll}
\author{Iain Matthews}
\affiliation{
\institution{Epic Games}
\country{USA}
}

\begin{abstract}
In recent years, the development of Neural Radiance Fields has enabled a previously unseen level of photo-realistic 3D reconstruction of scenes and objects from multi-view camera data.
However, previous methods use an oversimplified pinhole camera model resulting in defocus blur being `baked' into the reconstructed radiance field.
We propose a modification to the ray casting that leverages the optics of lenses to enhance scene reconstruction in the presence of defocus blur.
This allows us to improve the quality of radiance field reconstructions from the measurements of a practical camera with finite aperture.
We show that the proposed model matches the defocus blur behavior of practical cameras more closely than pinhole models and other approximations of defocus blur models, particularly in the presence of partial occlusions.
This allows us to achieve sharper reconstructions, improving the PSNR on validation of all-in-focus images, on both synthetic and real datasets, by up to 3 dB.
\end{abstract}

\ccsdesc[500]{Computing methodologies~Rendering}
\ccsdesc[500]{Computing methodologies~Reconstruction}
\ccsdesc[100]{Computing methodologies~Neural networks}

%
%

\keywords{Computational Photography, Multi-View \&3D, Differentiable Rendering, Neural Rendering}

\begin{teaserfigure}
    \centering
    \includegraphics[width=0.33\textwidth,trim={7cm 2.7cm 4.6cm 2cm},clip]{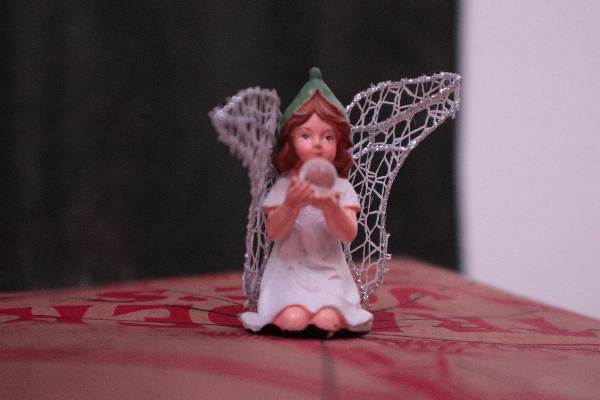}
    \includegraphics[width=0.33\textwidth,trim={7cm 2.7cm 4.6cm 2cm},clip]{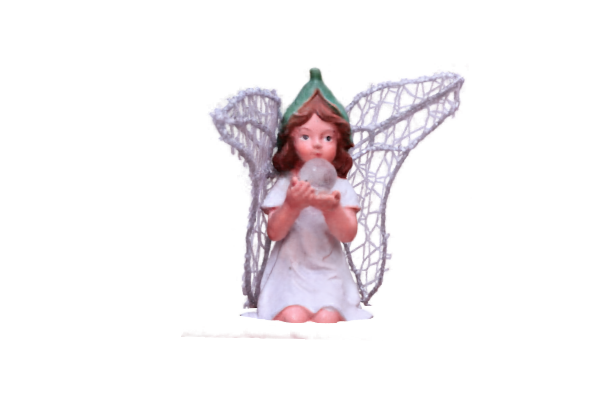}
    \includegraphics[width=0.33\textwidth,trim={7cm 2.7cm 4.6cm 2cm},clip]{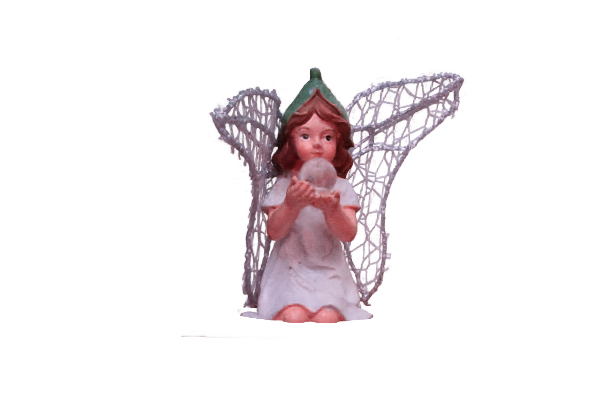}
    \caption{
    \textbf{Left:} a 5cm $\times$ 8cm $\times$ 6cm figurine captured with an ƒ/5.6 aperture, (\textbf{center}) reconstructed with instant NGP \cite{mueller2022instant}, and the proposed ƒNeRF (\textbf{right}).
    The reference image on the left was not used for reconstruction.
    While previous methods bake the defocus blur in to the radiance field, our method does not and reconstructs sharp details at all depths.
    It achieves this by successfully aggregating the radiance information from views close to the focus plane while simultaneously combining radiance values further away from the focus plane using a physically meaningful defocus model.
}
    \label{fig:wings}
\end{teaserfigure}

\maketitle

\input{intro.tex}
\input{related.tex}
\input{method.tex}

\input{implementation.tex}
\input{results.tex}

\input{discussion.tex}

{\small
\bibliographystyle{ACM-Reference-Format}
\bibliography{dof}
}

\end{document}

%% file: intro.tex
\section{Introduction}

Practical cameras are not pinhole cameras:
they use a lens with a finite aperture to focus light on the sensor, resulting in higher light throughput and advantages such as faster exposure time and less vignetting.
However, having a finite aperture inevitably results in defocus blur in the images.
This occurs because real camera pixels integrate photons coming from different parts of the aperture, instead of a single infinitesimally small pinhole.
Many modern radiance field reconstruction methods, such as \citet{mildenhall2021nerf}, use such data with traditional pinhole camera models under the assumption of a pixel-to-ray correspondence.
The resulting scene models have no choice but to `absorb' the blur in erroneous material or density models.

Since multi-view data includes multiple observations of the same scene point, the same point may be seen in sharp focus in some frames and be blurry in others.
If the forward imaging model closely matches the optics of a real camera lens, better scene reconstruction that preserves the sharp observation is possible.
Many existing methods approximate the physics of light rays and follow the original Neural Radiance Field paper~\cite{mildenhall2021nerf}, rendering pixel values by casting single rays from the pixel center through a pinhole.
MipNeRF~\cite{barron_mip-nerf_2021} and ZipNeRF~\cite{barron2023zipnerf} both propose sampling methods that are used to model a pixels integration of light from a cone; 
\citet{wu2022dof}  and \citet{pidhorskyi2022depth} find 2D convolution approximations for defocus. 
However, none of these methods account correctly for partial occlusions (depth discontinuities) and they do not model the lens aperture and focus plane in a physically meaningful way.
The ZipNeRF/MipNeRF cone aggregation strategies address sample aliasing, but incorrectly approximate partial occlusion within a pixel (for a visualization of the sample points on the cone surface, see Fig.~\ref{fig:samples}).
2D convolution approximations are lacking information that is invisible in the optical center, but would be visible from other parts of the aperture.
Methods for rendering physically-realistic camera measurements have been previously developed~\cite{pharr2023physically}.
Most prominently, one can use Monte-Carlo path tracing and trace multiple rays through the camera lens, refracting at the lens surface, to simulate the physical aggregation process of light at the sensor level.
Presumably, previous radiance field reconstruction methods have not adapted it because the volumetric rendering process was already considered slow and additional samples would further slow down the reconstruction, making it intractable.
Thanks to the recent progress in accelerating reconstruction and rendering for radiance fields~\cite{mueller2022instant,bozic2022neural,kerbl3Dgaussians}, these techniques become accessible---even more, we found that the practical runtime impact remains limited, presumably due to similarities in rays cast (for example, due to memory access patterns).
Hence, we propose ƒNeRF, a method that leverages the physics of real lenses and their light aggregation process to enhance scene reconstruction.

Our contribution lies in adapting classical rendering techniques for volumetric rendering with finite aperture cameras.
We show that this approach not only aligns more closely with optics, but also proves to be computationally tractable for the reconstruction of radiance fields.
We also demonstrate that, using analytical gradient for aperture size, both aperture size and focus distance can be jointly optimized using gradient descent.
This makes the proposed method easy to use in conjunction with many gradient-based 3D reconstruction techniques, such as neural radiance fields \cite{mildenhall2021nerf, mueller2022instant}, Gaussian splats \cite{kerbl3Dgaussians} or implicit surfaces \cite{wang2021neus}. 

\begin{figure}
    \centering
    \includegraphics[width=.48\textwidth]{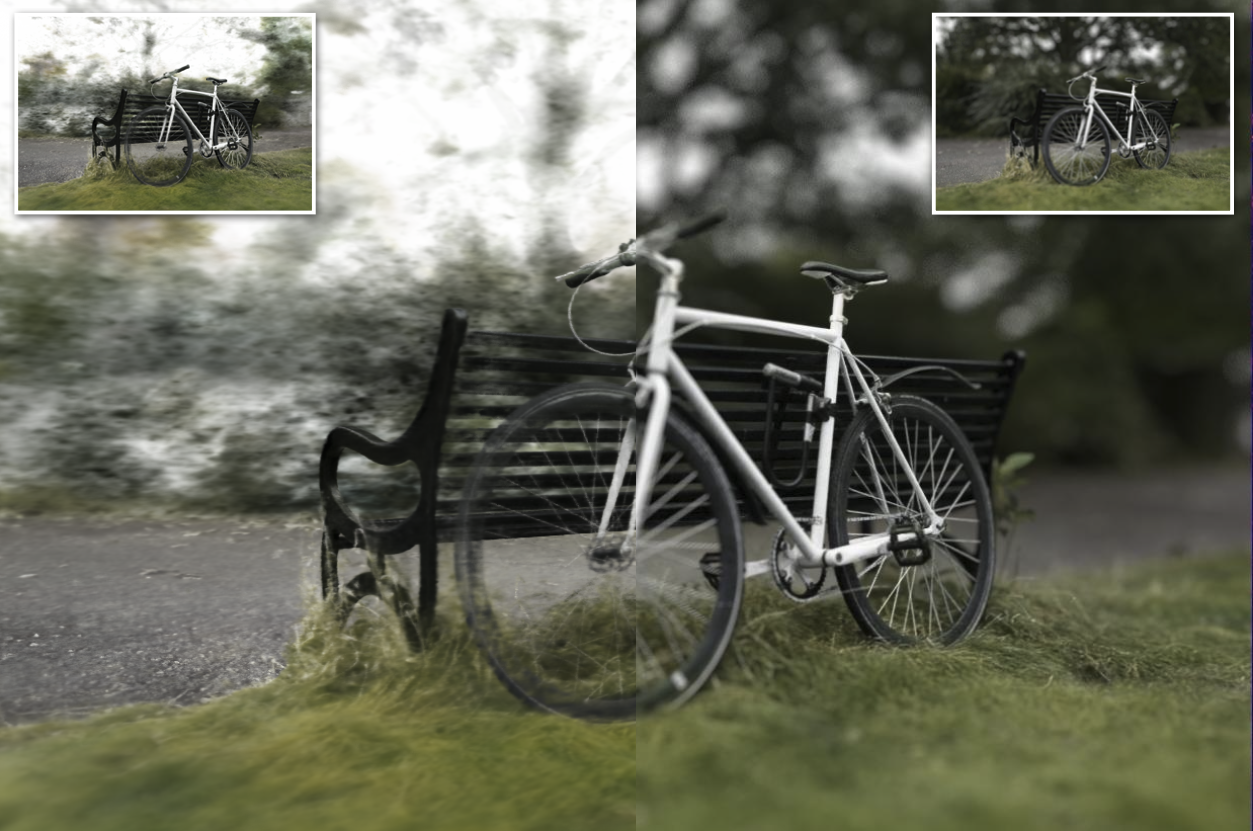}
    \vspace*{-0.7cm}
    \caption{Synthetic defocus on the MipNeRF360 `bicycle' scene~\cite{barron2022mipnerf360}. \textbf{Left} rendered with ZipNeRF~\cite{barron2023zipnerf} sampling cones modified to account for a larger aperture; \textbf{right} the proposed ƒNeRF.
    The modified ZipNerf forward model doesn't accurately model partial occlusions within a pixel, while ƒNeRF does and produces realistic blur and bokeh.}
    \label{fig:bicycle_defocused}
\end{figure}

In summary, we present a simple method for modeling lens defocus to significantly improve the sharpness of radiance field reconstruction from defocus blurred data.
We demonstrate the validity of our approach on synthetic and real data with defocus blur.

%% file: related.tex
\section{Related Work}
\thispagestyle{empty}

\subsection{Defocus blur in differentiable rendering}
Several existing methods for gradient-based 3D reconstruction address the depth of field effect.
\cite{mueller2022instant, mildenhall2021nerf} render synthetic defocus on radiance fields, but do not model defocus in the reconstruction process.
\cite{ma2022deblur, peng2023pdrf, Lee_2023_CVPR, lee2024sharp} reconstruct radiance fields from blurred input images by optimizing the ray positions without a physical lens model.
\cite{pidhorskyi2022depth, wu2022dof} approximate the effect as a convolution with depth-dependent blur kernel, and show that jointly optimizing the aperture size and focus distance results in sharper reconstructions than no defocus blur modeling. 
One practical consideration of the convolution-based approximations is that they require rendering of image patches proportional to blur size, instead of individual pixels, as well as careful handling of the convolution borders.
More importantly, the convolution approximation is inaccurate along object boundaries and degrades reconstruction quality. We found that the physically-realistic approach to modeling defocus blur results in better reconstructions, while our implementation keeps the reconstruction time within a reasonable range (Fig.~\ref{fig:nrays_exp}.). We show a comparison on defocused data from \cite{wu2022dof} in \ref{para:dofnerf}\footnote{We contacted \citeauthor{pidhorskyi2022depth} but were not able to get access to their data or code for establishing a comparison.}.

\subsection{Camera modeling}
Many papers show that better camera modeling improves the reconstruction of NeRFs from real data.
\cite{park2023camp, Lin_2021_ICCV, wang2021nerfmm} jointly optimize camera intrinsics and extrinsics while
\cite{SCNeRF2021, Xian_2023_CVPR} focus on the optimization of lens distortion.
Since all these methods only cast a single ray from each pixel, they can all be easily extended with the proposed method for modeling finite aperture of practical cameras.

Inspired by the success of NeRF, recent camera modeling works are embracing the use of implicit networks with coordinate inputs.
Some use a network to predict the blur at each pixel and use it to render the camera measurements \cite{lin2023learning},
and demonstrate improvements on multiple reconstruction problems, including reconstructing all-in-focus image from image stacks \cite{wang2023neural, huang2023inverting}.
These previous works capture a scene with different configurations of the camera, such as changing the exposure time or focus distance, while we take a camera with fixed configuration and focus on reconstructing a high quality radiance field of the scene.
Additionally, these works are less physically grounded as they do not explicitly model ray casting from the aperture, so they can not optimize the aperture radius. In this paper we propose an efficient method for rendering the gradient of aperture radius.

\subsection{Sampling volumes in NeRF}
Multiple works \cite{barron_mip-nerf_2021, barron2023zipnerf} address the aliasing problem in rendering radiance fields.
The need for anti-aliasing comes from the fact that each pixel measurement integrates photons across the pixel area as well as across the aperture. 
However, the cone modeling in MipNeRF and ZipNeRF, from which the radiance field is sampled, only accounts for sampling the pixel area and does not take into account the camera aperture.

To incorporate aperture sampling two modifications are needed.
First, the apex of the cone should not be the pinhole lens position, but it should rather lie on the focus plane. 
Lens optics refract light passing the same point on the focus plane to the same pixel on the sensor plane.
Second, the width of the cone should also be expanded so that it matches the aperture size (see Fig.~\ref{fig:samples}).
We evaluate the effectiveness of this modified version of ZipNeRF with correct focus point location and cone width matching the aperture, and compare its performance against the other models in Fig.~\ref{fig:synth_lego}, Table~\ref{tab:synth_ssim} and Table~\ref{tab:synth_psnr}.
The result shows ZipNeRF with aperture cone modifications results in sharper details in comparison to the original ZipNeRF.

MipNeRF and ZipNeRF assume that all points in their sampled Gaussian share the same occlusion.
This approximation allows for fast evaluation by allowing them to integrate in encoding space and therefore limiting the number of queries to the MLP. This is acceptable for approximating the sampling of the pixel area, however for larger apertures it is inaccurate for partial occlusions within a projection cone, which often occur at object boundaries.
The difference is shown in Fig.~\ref{fig:bicycle_defocused}.
We show that with a more physically accurate model we can obtain better reconstructions for large apertures, in synthetic (Fig.~\ref{fig:synth_lego}) and real data (Fig.~\ref{fig:mipnerf360}).

%% file: method.tex
\section{Method}
\thispagestyle{empty}
In this section, we start by reviewing the mathematics of defocus blur. 
We then derive our method which extends volumetric rendering with aperture modeling, which allows us to reconstruct sharp radiance fields from defocus blurred measurements. 
For joint estimation of the defocus blur parameters with the radiance field, we also derive a method to efficiently render aperture radius gradient for circular apertures.

\begin{figure}
  \centering
   \includegraphics[width=\linewidth]{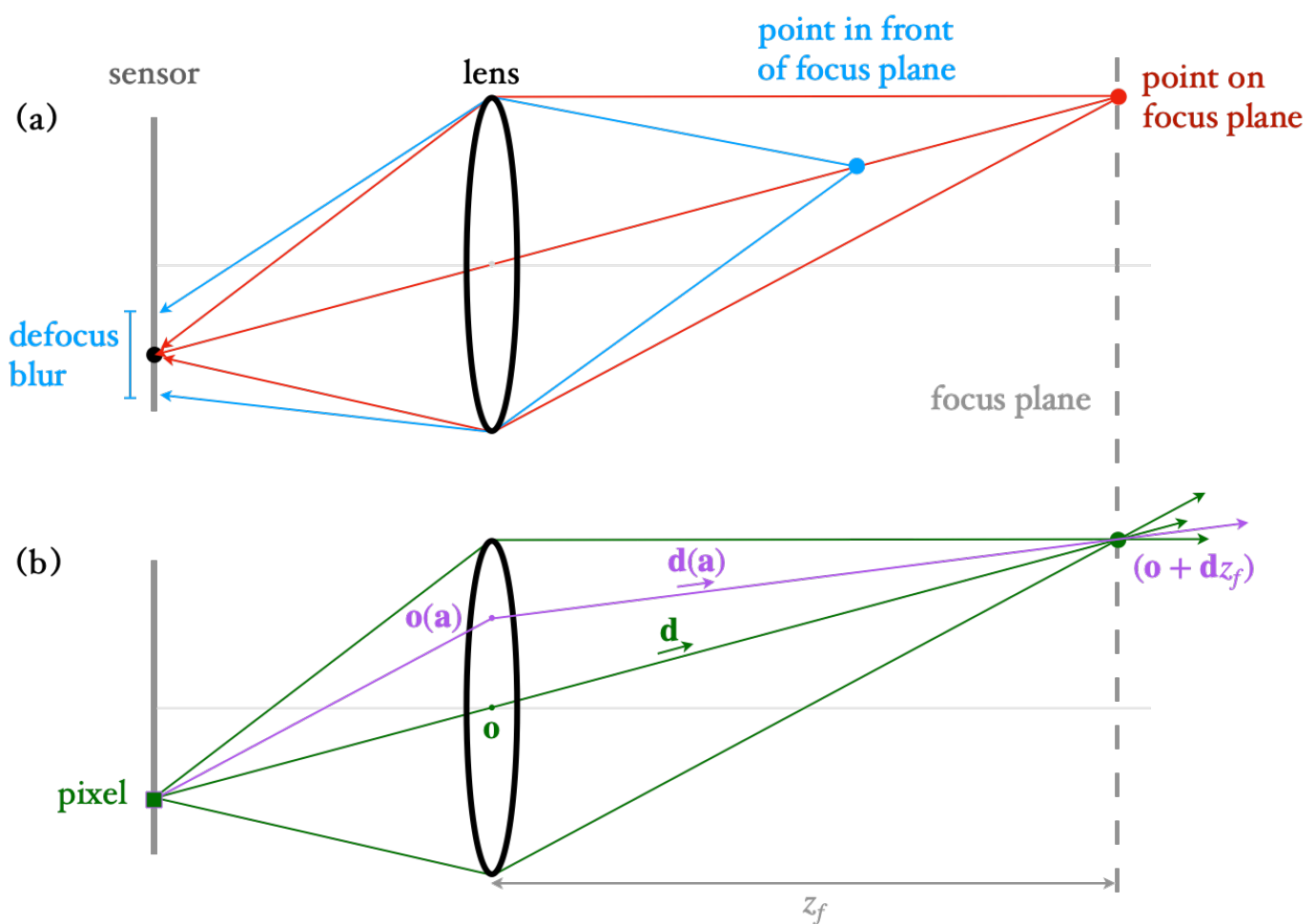}
   \caption{
   Top: The red point on the focus plane results in a sharp image, while the blue point in front of the focus plane results in a blurry image, or bokeh.
   Bottom: to render color at pixel $\bfp$, we draw samples $\bfa$ from the aperture and cast modified rays $(\bfo(\bfa), \bfd(\bfa))$ from it.
  }
   \label{fig:method}
\end{figure}

\subsection{Defocus blur}
Pinhole cameras have limited light throughput and suffer from heavy vignetting effects. In practice, cameras use a lens with a larger aperture to capture more light \cite{Allen:2011}.
This allows either capturing with a shorter exposure time (important for moving subjects) or in dimly lit environments.
However, a larger aperture creates defocus blur and hence leads to a loss of detail in areas further away from the focus plane \cite{potmesil1981synthetic}.
Fig.~\ref{fig:method}(a) shows how the image of two points is formed on a camera with a finite aperture.
For rays emitted by the red point on the focus plane, the lens refracts them so that all of them converge to the same point on the sensor, forming a sharp image of the point.
However, for rays emitted by the the blue point in front of the focus plane, the lens is unable to focus them and each ray will intersect the sensor at a slightly different point, spreading out its image, an effect known as bokeh. 

For a camera with aperture radius $a_R$, focal length $f$ and focused at distance $z_f$, the radius of defocus blur, often known as circle of confusion, for scene point at depth $z$ is,
\begin{equation}
a_R \frac{\vert z - z_f \vert}{z}\frac{f}{z_f-f}.
\end{equation} 
%
The defocus blur is the same for scene points at the same depth, thus previous methods \cite{wu2022dof, pidhorskyi2022depth} approximate the defocus blur effect as 2D convolutions between a sharp image of the scene and a bokeh kernel. 
However, using 2D convolution approximations leads to inaccuracies at occlusion boundaries and requires discretizing a 3D scenes into depth layers.
Volumetric rendering as used by many radiance field rendering methods offers us an opportunity to render higher fidelity defocus blur avoiding these problems.

\subsection{Volumetric rendering with finite aperture}
Most previous volumetric rendering methods model all cameras as pinhole cameras and only generate rays from the center of a pixel to the pinhole point.
Specifically, the original NeRF \cite{mildenhall2021nerf} expects color $\widetilde{C}$ from ray $\bfr$ from a scene with volume density $\sigma(\bfx)$ and emitted color $\bfc$,
\begin{equation}
    \widetilde{C}(\bfr) = \int_{t_n}^{t_f} T\left(t\right) \sigma\left(\bfr(t)\right) 
    \bfc\left(\bfr(t), \bfd\right) dt,
\end{equation}
where $T(t) = \exp{-\int_{t_n}^t \sigma\left(\bfr(s)\right) ds}$, and $t_n$, $t_f$ are respectively the near and far depth that bounds the scene.
Modeling the lens aperture expects pixel color $C(\bfp)$ integrated over many rays across different parts of the aperture $\mathcal{A}$. 
We follow classical rendering \cite{pharr2023physically} and sample rays from the aperture,
\begin{equation}
    C(\bfp) = \frac{1}{ \text{area}\left(\mathcal{A}\right) } \int_{\bfa \in \mathcal{A}} \widetilde{C}\left(\bfr (\bfa) \right) d\bfa,
    \label{eq:forward}
\end{equation}
where $\bfr(\bfa)$ modifies a ray $\bfr = (\bfo, \bfd)$ with origin $\bfo$ at the pinhole and direction $\bfd$ for finite aperture width.
For focus distance of $z_f$, the modified ray $\bfr(\bfa) = (\bfo(\bfa)), \bfd(\bfa))$ has its origin on the aperture 
\begin{equation}
\bfo(\bfa) = \bfa
\end{equation}
 and direction $\bfd(\bfa)$ pointing to the focus point, 
\begin{equation}
\bfd(\bfa) = \frac{\bfo + \bfd z_f  - \bfa }{\Vert \bfo + \bfd z_f  - \bfa \Vert^2_2}.
\end{equation} 
See Fig.~\ref{fig:method} (b).

Our implementation samples points $\bfa$ from the aperture using a quasi-Monte Carlo approach using a Sobol Sequence~\cite{sobol1967distribution}. This improves reconstruction performance, especially at low sample counts.
If each pixel samples multiple rays, we add offsets to the points sampled during ray matching to reduce z-aliasing.
For $i$-th ray among the $n$ rays we sample for each pixel, we sample the point in ray section $(t_{start}, t_{end})$ at,
\begin{equation}
    t=\frac{i}{n} t_{start} + \left(1-\frac{i}{n}\right) t_{end}.
\end{equation}

\subsection{Optimizing aperture and focus depth}
Highest quality reconstructions require accurate estimates of the camera parameters.
Structure-from-motion packages together with EXIF data from recordings or manual recording of the parameters can provide great starting points.
The dense, pixel-wise optimization of the entire scene provides ample opportunity to improve over these initializations and leads to better results~\cite{Xian_2023_CVPR,park2023camp}.

Our extended model has additional parameters for focus distance and aperture radius.
The focus distance can naturally be optimized by using the auto-differentiation of the ray casting operation.
The aperture radius, however, is a parameter for the sampling operation and its auto-differentiation gradient has high variance, due to the random sampling.

We follow \cite{pidhorskyi2022depth} and render the gradient of aperture radius from its analytic gradient. The gradient of pixel color $C(\bfp)$ with respect to aperture radius $a_R$ as,
\begin{align}
    \frac{d C(\bfp) }{d a_R} 
    &= \frac{2}{a_R} C_{\text{ring}}(a_R) + 0 - \frac{2}{a_R} \int_{a_r=0}^{a_R} \frac{2a_r}{a_R^2} C_{\text{ring}}(a_r) da_r\\&
    = \frac{2}{a_R} \left( C_{\text{ring}}(a_R) - C(\bfp)\right) \label{eq:aperture_grad}.
\end{align}

Thus, we can render $C_{\text{ring}}(a_R)$ by sampling points on the boundary of the aperture and use Eq.~\eqref{eq:aperture_grad} for computing the gradient of the aperture radius; this gives us stable convergence of the aperture radius, as shown in Table~\ref{tab:synth_opt_dof_ssim}.

%% file: implementation.tex
\section{Implementation}

We implement our framework using neural graphics primitives for fast reconstruction and rendering speeds and a compact volume representation~\cite{mueller2022instant}. 
Additionally, we use a dense volume acceleration structure for empty space skipping~\cite{li2023nerfacc}.
Similar to NerfAcc, we use a sample-point target to guide ray batching and adjust the number of pixels sampled per batch dynamically.
We use an ADAM optimizer~\cite{adam2014} with an exponential learning rate reduction schedule with two steps to guide the optimization towards convergence.
We observed that it is critical to average colors in linear space to create realistic bokeh effects, in particular in bright areas, and account for the logarithmic transfer function in sRGB encoded images during our reconstruction.
We use the same batching, number of steps and learning rate schedule across all methods for a fair comparison.
The experiments are run on NVIDIA A100 GPUs with 80GB memory unless otherwise specified.

%% file: results.tex
\section{Experiments}
\thispagestyle{empty}
\subsection{Synthetic experiments}
\subsubsection{Experiment setup}
We validate our method on the synthetic data rendered with large aperture; see Fig.~\ref{fig:synth_lego}.
We render the blender scenes from \cite{mildenhall2021nerf} in BlenderNeRF\footnote{Blender computes ƒ numbers differently from photography conventions; to reproduce the rendering use ƒ/0.2 and focal length 50mm camera in Blender.} \cite{Raafat_BlenderNeRF_2023} with \textflorin/5 and focus distance 3.5m for 100 images captured from a hemisphere with radius 4m. 
We render the same images sharp and without defocus for testing. 

To isolate the effect of addressing defocus, we only use the reconstruction loss between measured pixel color $C$ and rendered pixel color $C(\bfp)$, 
\begin{equation}
    \ell_{recon} = \text{smooth }\ell_1\left(C(\bfp) - C \right),
\end{equation}
where smooth $\ell_1$ loss equals $\Vert x \Vert^2_2$ for $\vert x\vert <= 1$ 
 and $\vert x \vert$ otherwise; we use it for robust optimization. 
We use the ground truth camera intrinsics, extrinsics, aperture radius, and focus distance unless otherwise specified.
All experiments are run with learning rate 1e-2 for a fixed number of 2e4 steps. 
We use base resolution of 16 with 16 levels and a maximum resolution of 4096 for the NGP volume, backed by a hashmap of size $2^{19}$.
The target number of point samples per batch is 262,144 for all methods.

\subsubsection{Comparison against ZipNeRF and ZipNeRF modified for aperture.}
\begin{figure}
    \centering
    \includegraphics[width=.9\linewidth]{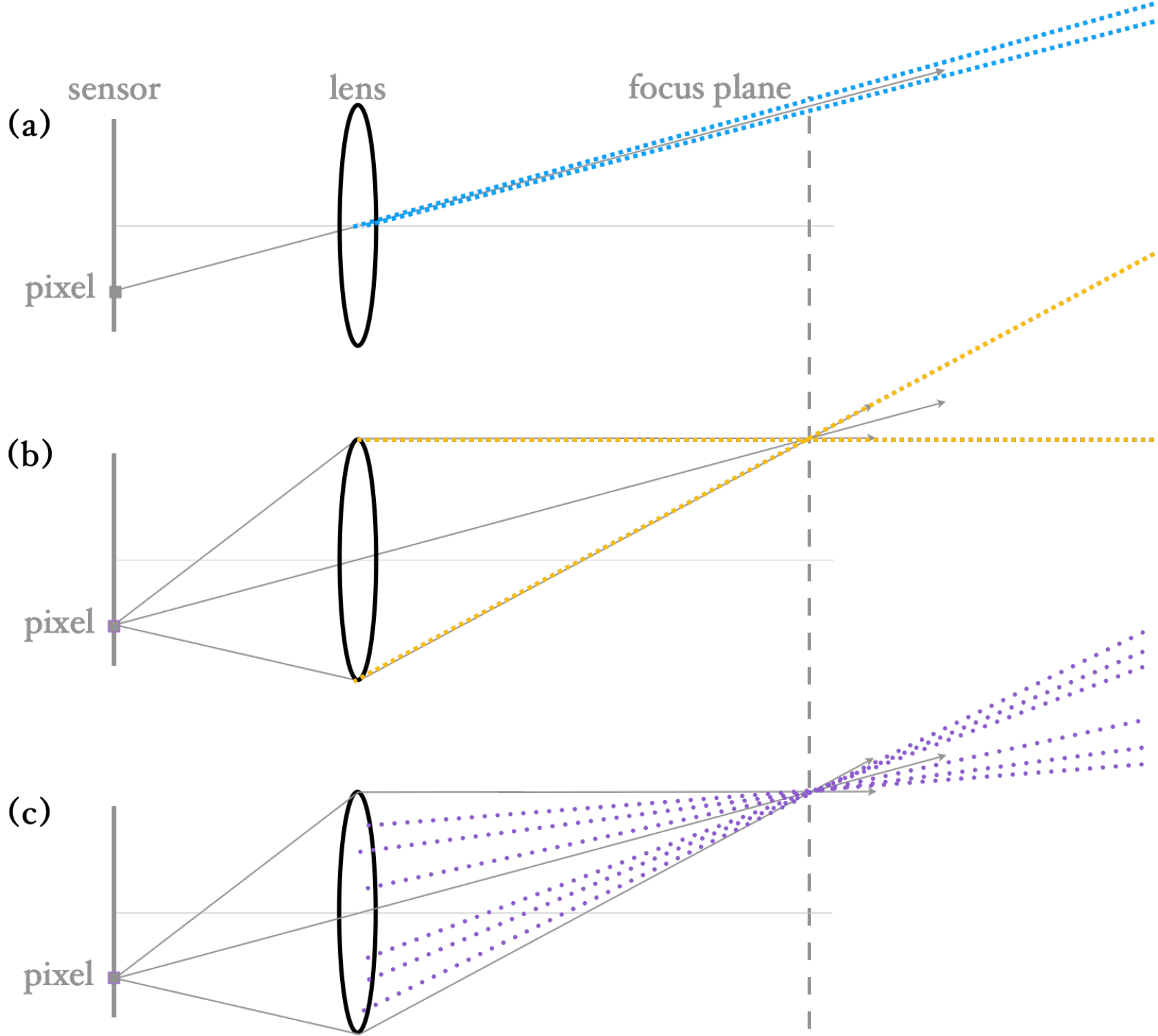}
    \caption{Schematic of sampling locations of different methods in 2D. 
    (a) blue points show ZipNeRF samples on a small cone with apex on the pinhole location;
    (b) yellow points show ZipNeRF modified for aperture by moving the apex to focus plane and expanding the cone to match aperture;
    (c) purple points show ƒNeRF samples on 6 rays, drawn at random, passing through the aperture.
    ZipNeRF sample solely on a cone surface whereas the proposed method casts rays within the cone volume.
    We show actual sampling locations in 3D in the supplementary video.
    }
    \label{fig:samples}
\end{figure}

We implemented the multisampling and downweighting from ZipNeRF (ZN), as well as a modified version where we shift the sampling cone apex to the focus plane and expand the cone to match the aperture (ZN+A) for comparison with the proposed ƒNeRF method.
The modification in the cone shape is shown in the sampling locations in Fig.~\ref{fig:samples}.
We report the SSIM and PSNR metrics on the sharp testing images in Table \ref{tab:synth_opt_dof_ssim}.
An Instant NGP (NGP) reconstruction with pinhole camera model does not model lens aperture and bakes in the defocus blur in the scene, as shown on second column.
The ZipNeRF (ZN) reconstruction models a cone defined by the pixel, not the lens aperture, and fails on large aperture scenarios as shown.
ZipNeRF with the sampling cone modified for aperture (ZN+A) produces sharper results, but the model mismatch in partial occlusions
tends to produce over-smoothed rendering and an over-sharpened reconstruction. 
ZN+A fails to reconstruct the drum scene due to the large amount of partial occlusions and specularity in this scene.
Our proposed method with aperture modeling significantly improve the sharpness of reconstruction, with 32 rays (LN-32) showing more details than 6 rays (LN-6) on most scenes, as shown on the last two columns.

\input{fig_synth_cmp.tex}

\subsubsection{Number of rays per pixel.}
\begin{figure}[tb]
    \centering
    \includegraphics[width=\linewidth]{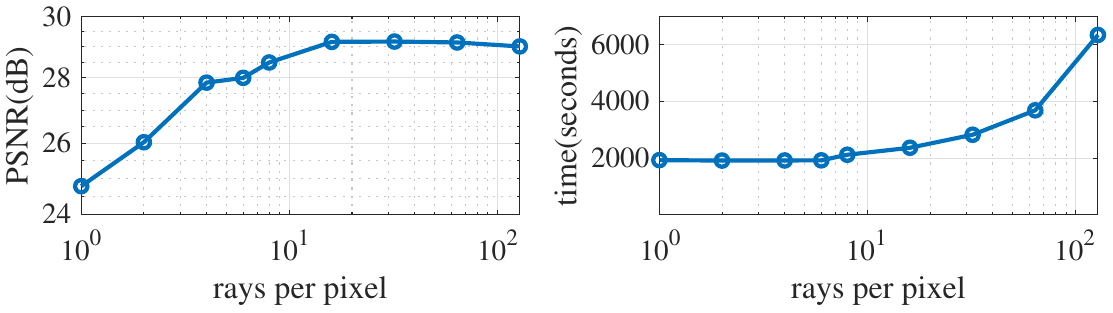}
    \caption{Reconstruction quality and runtime v.s. number of rays per pixel on synthetic lego data.
    The reconstruction quality saturates near 16 rays per pixel; the runtime does not decrease significantly below 8 rays per pixel. 
    }
    \label{fig:nrays_exp}
\end{figure}

To understand how the reconstruction quality and runtime scale with the number of rays we sample for each pixel, we reconstruct the synthetic Lego scene with the proposed ƒNeRF method with different numbers of rays per pixel, but with a constant number of pixels per batch (1024) Fig.~\ref{fig:nrays_exp}.
This experiment is run on a NVIDIA Quadro RTX 8000.
The reconstruction quality saturates near 16 rays per pixel.
We also see that the runtime does not increase significantly below 8 rays per pixel; we posit that the lower increase in runtime originates from synergy effects of the many rays in each batch being closely related in terms of memory accesses and computation. 

\subsubsection{Optimizing aperture radius and focus distance.}
To evaluate the convergence of the aperture radius and focus distance parameters, we reconstruct the synthetic Lego scene with mismatched initial parameters ($80\%$, $100\%$, $120\%$).
As shown in Tab.~\ref{tab:synth_opt_dof_ssim} 
mismatched defocus parameters result in degraded reconstruction quality, in some cases leading to complete failures to reconstruct the scene.
Allowing for joint optimization of the defocus parameters with the scene leads to comparable quality with correctly optimized defocus parameters up to a range of 20\% of the aperture and focus depth initialization.
This means that in practice, the proposed method can be expected to work reliably even for ballpark estimates of these parameters.
\thispagestyle{empty}

\begin{table}[tb]
\centering
\resizebox{0.48\textwidth}{!}{
\begin{tabular}{cc|cc}     
aperture init. &	focus init. & scene only & + opt. defocus params.\\
\hline
$80\%$ & $100\%$ & 28.73 | 0.7606 & \textbf{29.19 | 0.9295}\\
\underline{$100\%$} & \underline{$100\%$} & 28.98 | 0.9294 & \textbf{29.07 | 0.9311} \\
$120\%$ &$100\%$ & 27.72 | 0.8364 & \textbf{29.13 | 0.9276} \\
$100\%$ & $80\%$ & 18.55 | 0.7385 & \textbf{24.14 | 0.9341} \\
\underline{$100\%$} & \underline{$100\%$} & \textbf{29.04 | 0.9368} & 29.00 | 0.9366 \\
$100\%$ &$120\%$ & 22.86 | 0.8199 & \textbf{28.97 | 0.9338} 
\end{tabular}
}
\caption{Reconstruction PSNR|SSIM from different aperture radius and focus distance initializations for the synthetic lego scene, for scene only and joint optimization with defocus parameters. The aperture and focus initialization is perturbed as specified in the leftmost columns. When optimizing the defocus parameters we manage to maintain near perfect reconstruction quality. Note that there is a slight variance due to the stochastic optimization in our method comparing the two lines initialized with ground truth.}
\label{tab:synth_opt_dof_ssim}
\end{table}

\subsection{Experiments on Recorded Data}
\thispagestyle{empty}

\subsubsection{Experiment setup.}
To evaluate reconstruction on recorded data, we optimize camera poses using CamP~\cite{park2023camp} (using SE3, sampling 128 points in the reconstruction volume, with diagonal weight $\lambda=$1e-1, and identity weight $\mu=$1e-8 ) and the distortion loss from \cite{barron2022mipnerf360} ($\lambda_{dist} = $1.5938e-3). 

\begin{figure}[tb]
    \centering
    \includegraphics[width=0.48\textwidth]{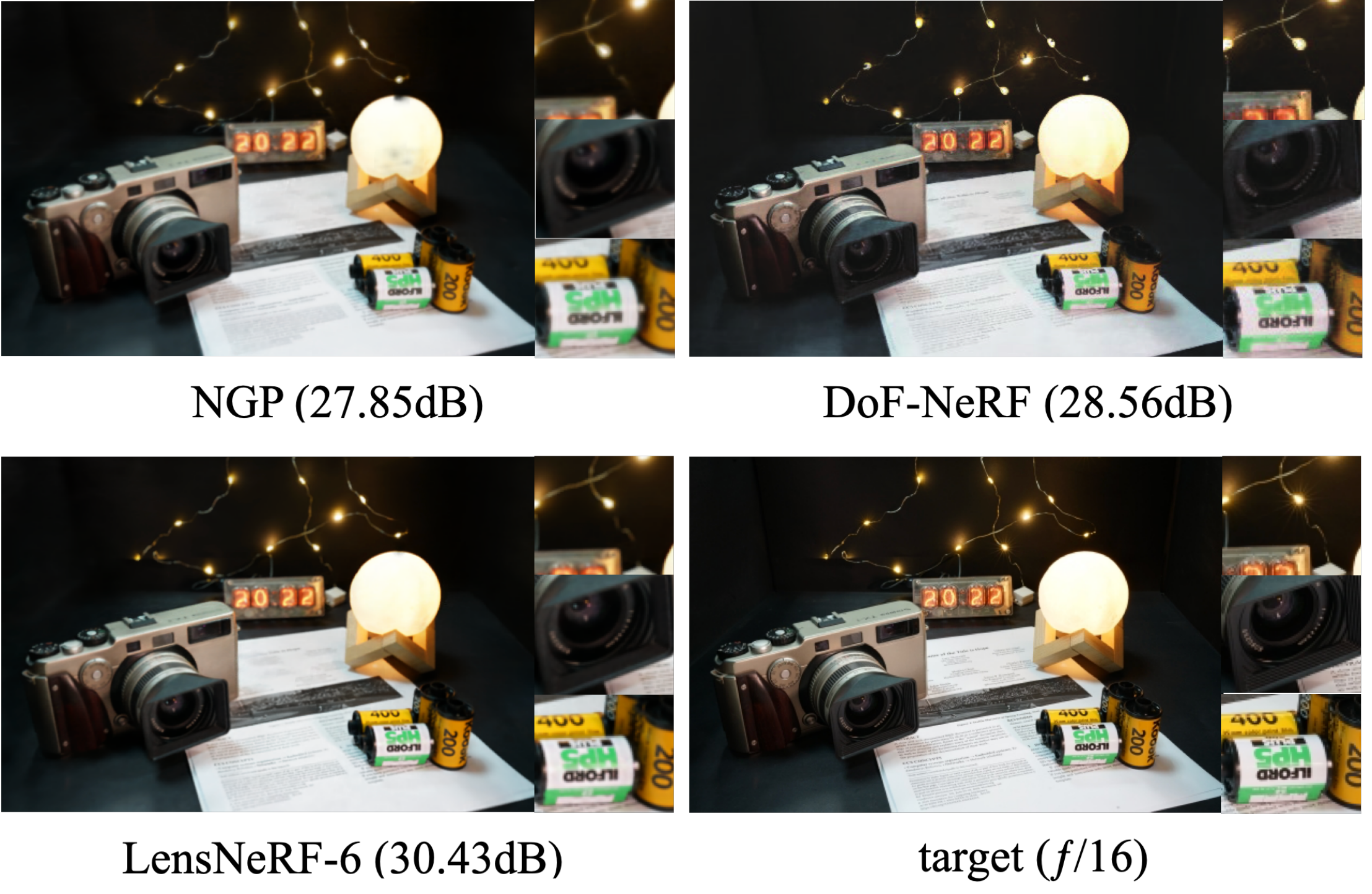}
    \caption{
    DoF-NeRF dataset reconstructions from (\textflorin/4) captures. 
    The numbers are PSNRs evaluated against \textflorin/16 images at 2 unseen poses.
    }
    \label{fig:cmp_dof_nerf}
\end{figure}

\subsubsection{DoF NeRF experiments.}
\label{para:dofnerf}
We validate our method on data from DoF-NeRF~\cite{wu2022dof}, which captures scenes with different focus distances and different apertures (\textflorin/4 and \textflorin/16).
We show comparisons with DoF-NeRF in Fig.~\ref{fig:cmp_dof_nerf}.
The input consists of images captured with 2 focus distances with the same camera pose, each consisting of 22 images.
We run our instant NGP implementation with pinhole camera model as a baseline.
For DoF-NeRF, since their code is unavailable, we use the PSNR number and result presented in their paper.
We follow the configuration described in the DoF-NeRF paper and reconstruct with ƒNeRF at the same resolution, with input dimension of $497 \times 322$, and jointly optimize aperture radius and focus distances.
However, unlike DoF-NeRF which estimates the per-frame  defocus

\subsubsection{Mip-NeRF360 experiments}
\begin{figure*}[p]
    \centering
    \includegraphics[width=\textwidth]{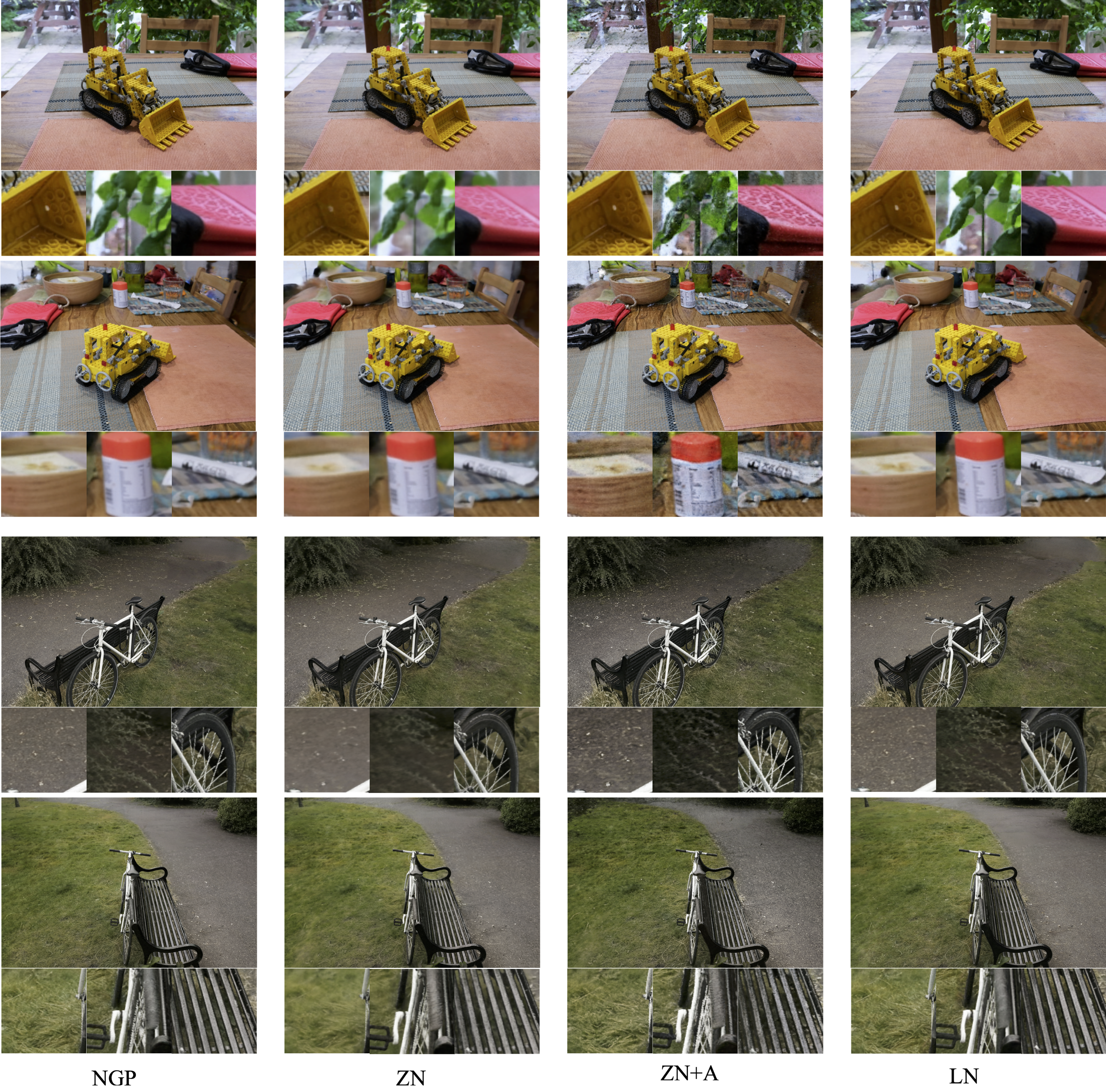}
    \caption{Results on MipNerf360 \cite{barron2022mipnerf360} data. Both NGP an ZipNeRF bake defocus blur into their reconstructions, resulting in smoothed out details. ZN+A generates over-sharpened details and introduces artifacts near occlusion boundaries. ƒNeRF reconstructs better details without artifacts.
    Please refer to our supplementary results for a better visualization of the reconstructions.
    }
    \label{fig:mipnerf360}
\end{figure*}

We also evaluate our method on the Mip-NeRF360 dataset.
While the apertures are fairly small, we can still improve the results and produce sharper details in the reconstruction in Fig.~\ref{fig:mipnerf360}.
Since the captures are centered around an object, most background points were similarly defocused across all frames.
The quantitative validation metrics are similar to baseline due to the validation set sharing the same defocus blur as input.
Since each scene contains more than a hundred images at high resolution, we find a balance between pixels per batch and rays per pixel, by starting at randomly sampling 1 ray per pixel and doubling it for each epoch.
The kitchen scene is reconstructed with 5e-4 aperture radius and 1.09 focus distance, and the bicycle scene with 1.5e-4 aperture radius and 0.15 focus distance.  
We use three sets of hyperparameters across all experiments in the paper: we adjust them between synthetic data, indoor and outdoor spaces.
To provide a rough overview of their ranges: the max resolution of the NGP is between 4096 and 467830, backed by a hashmap of size $2^{19}$ - $2^{21}$.
The overall learning rate for the ADAM optimizer is chosen between 0.01 and 0.0001, pose between 1e-5 and 1e-4, and intrinsics between 4e-6 and 8e-6.
We also regularize the deviation from initial camera pose, with a rotation loss using the $\ell_2$ distances between initial and optimized rotation matrices weighted by $\lambda_{\text{rot}}$=9.59e-3, and a translation loss using the $\ell_2$ distance between initial and optimized camera locations weighted by $\lambda_{\text{trans}}$=9.09e-3.
The learning rate schedule is being applied with two reduction steps with a size between 0.2 and 0.4.
Other parameters include weight decay (2e-6 to 9e-6) and gradient scaling (600x-6000x) for better numerical stability for compatibility with FP16 tensor operations.

\subsubsection{New data.}
We image a miniature figurine of size 5cm$\times$ 8cm $\times$ 6cm with a Canon EOS M100 with \textflorin/5.6.
Since the subject is small, we capture it with the camera handheld close to it, resulting in a close focus distance and a shallow depth of field.
We captured 96 images spread around the upper hemisphere.
We reconstruct using 88 frames of 10 times downsampled images, at $600\times400$ resolution.
One of the practical problems of reconstructing from defocus blurred data is that the initial camera poses from structure from motion are often inaccurate.
This is not a problem if the images are from a pre-calibrated multi-camera capture system.
DoF-NeRF solves this problem by capturing with both large and small aperture from the same positions and using the poses estimated only from small aperture photos.
We do not capture any photos at small apertures and found that we can get reasonable pose estimates using Agisoft Metashape\footnote{\url{https://www.agisoft.com}} using downsampled data.
This results in high quality reconstructions that we can segment in 3D.
You can find a held out image, as well as reconstruction renders in Fig.~\ref{fig:wings}.

%% file: fig_synth_cmp.tex
\begin{figure*}[tb]
    \centering
    \includegraphics[width=0.99\textwidth]{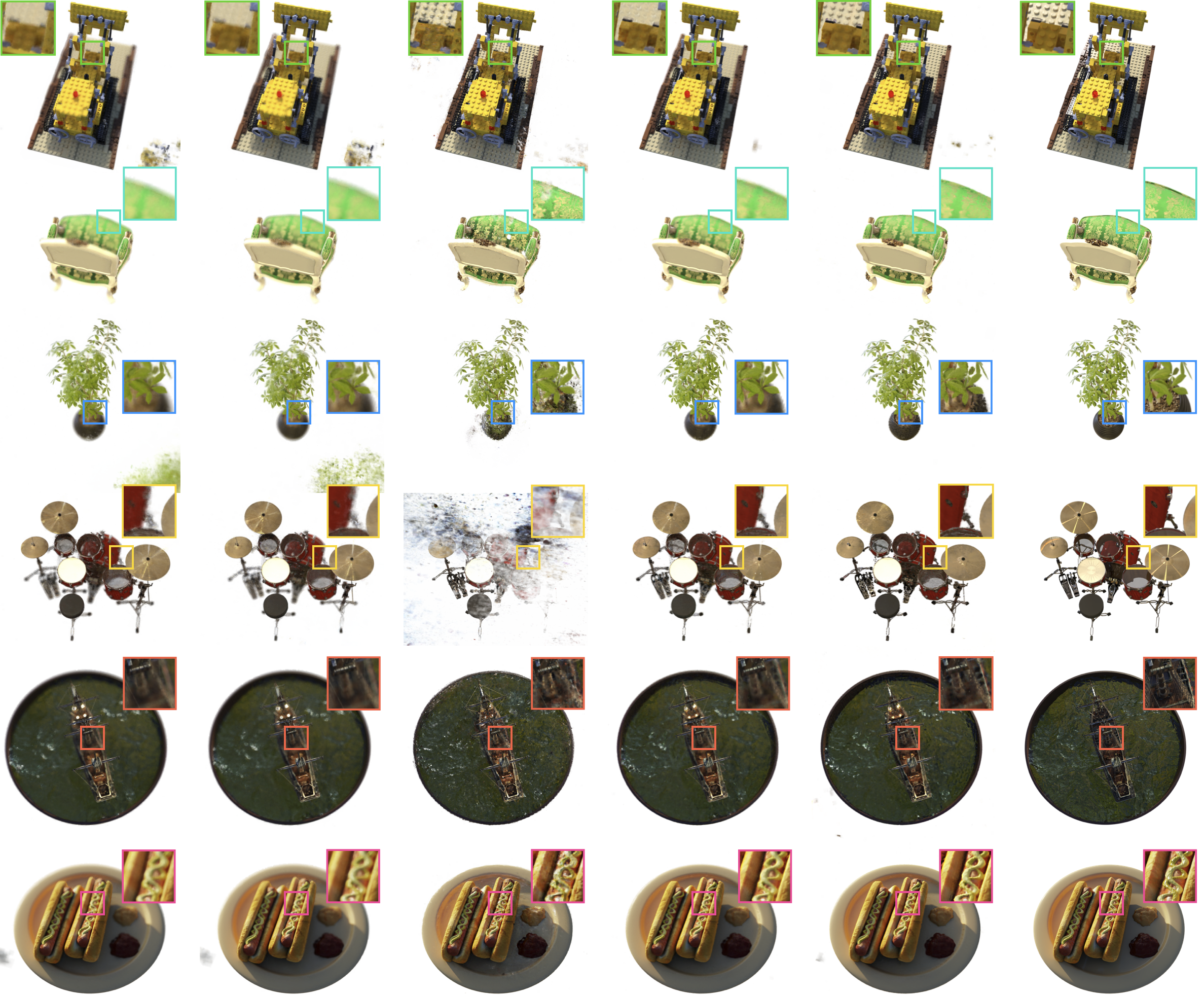}
    \caption{
        Synthetic data results.
        From left to right: input frame closest to test viewpoint, reconstruction from iNGP, ZipNeRF, ZipNeRF modified for aperture, the proposed method LensNeRF with 6 rays per pixel, LensNeRF with 32 rays per pixel, and all-in-focus ground truth.
        The closest input views demonstrate notably blurry regions in important areas of the image, and our reconstructed model is not able to leverage this viewpoint for the reconstruction of this viewing angle.
        However, other viewpoints were sufficient to reconstruct the areas in question with high fidelity and sharpness and the model is not negatively affected by the reconstruction loss thanks to our accurate depth-of-field model.
    }
    \label{fig:synth_lego}
\end{figure*}

\begin{table}[tb]
\centering
\begin{tabular}{c|ccccc}
scene &	NGP & ZN & ZN + A & LN 6 & LN 32\\
\hline
lego    & 	0.8801& 	0.8806& 	0.8800& 	0.9214& 	\textbf{0.9360}\\
chair   & 	0.9023& 	0.9023& 	0.9303& 	0.9261& 	\textbf{0.9419}\\
ficus   & 	0.9300& 	0.9395& 	0.9361& 	0.9645& 	\textbf{0.9657}\\
drums   & 	0.8826& 	0.8838& 	0.6085& 	\textbf{0.9037}& 	0.9033\\
mic     & 	0.9157& 	0.9157& 	0.9299& 	0.9325& 	\textbf{0.9377}\\
ship    & 	0.7352& 	0.7362& 	0.7017& 	\textbf{0.7515}& 	0.7467\\
hotdog  & 	0.9373& 	0.9377& 	0.9361& 	0.9549& 	\textbf{0.9572}
\end{tabular}
\caption{Reconstruction SSIM from synthetic defocused measurements with different methods.}
\label{tab:synth_ssim}
\end{table}

\begin{table}[tb]
\centering
\begin{tabular}{c|ccccc}     
scene &	NGP & ZN & ZN + A & LN 6 & LN 32\\
\hline
lego	&   25.75&  25.84& 	26.02& 	       28.26& 	        \textbf{29.00}\\
chair   &	27.25& 	27.24& 	27.84& 	      28.9& 	        \textbf{29.64}\\
ficus   &	27.28& 	27.65& 	26.32& 	      \textbf{30.65}& 	30.49\\
drums   &	22.54& 	22.46& 	14.53& 	      \textbf{23.93}& 	23.43\\
mic     &	24.18& 	24.17& 	\textbf{26.10}& 	25.84& 	    26.07\\
ship    &	22.64& 	22.68& 	21.46& 	      \textbf{23.51}& 	23.17 \\
hotdog  &	29.63& 	29.69& 	30.06& 	       31.93& 	        \textbf{32.08}
\end{tabular}
\caption{Reconstruction PSNR from synthetic defocused measurements with different methods.}
\label{tab:synth_psnr}
\end{table}

%% file: discussion.tex
\section{Conclusion}
\thispagestyle{empty}

In this paper, we improved the camera modeling in radiance field methods and pushed them closer to the realities of practical camera systems.
The proposed method, LensNeRF, can be easily integrated into most radiance field reconstruction systems.
It casts multiple rays from the camera aperture to render physically realistic defocus blur.
We show that LensNeRF reconstructs sharp radiance fields and is computationally tractable.
By incorporating the optics of a thin lens into our modeling, we are able to render the analytic gradient of aperture radius. We are successful at jointly estimating the aperture radius and focus distance, two parameters that control the amount of defocus, along with the scene.
We have critically evaluated the current state of art volume sampling approach, ZipNeRF, and illustrated the effects of approximations included in that approach. We introduce an aperture sampling extension to ZipNeRF to more reasonably compare with our LensNeRF approach in scenarios involving partial occlusions.
Our experiments on synthetic and real defocused camera inputs show up to 3dB PSNR improvement when evaluated on all-in-focus novel views.

While this work represents a stride toward more accurate modeling of practical camera systems, it is important to acknowledge its limitations.
One of the key simplification is the use of the thin lens model, which, although effective, does not encapsulate the complexity of modern camera lenses with multiple elements.
Some unique characteristics of cameras, such as chromatic aberration, lens flare, and bokeh variation cross different regions of the lens, are not represented in our model.
Further research could focus on integrating more expressive camera models that capture these nuances while maintaining rendering efficiency.
These advancements will push the fidelity of reconstructions from practical camera inputs.